\pdfoutput=1
\documentclass{article}
\usepackage{spconf,amsmath,epsfig}

\usepackage{amsmath,amssymb,amsfonts}
\usepackage{algorithmic}
\usepackage{graphicx}
\usepackage{textcomp}
\usepackage{xcolor}
\usepackage{bbm}
\usepackage{multirow}
\usepackage{titlesec}
\titlespacing*{\section}{0pt}{1.4ex plus .3ex minus .2ex}{0.65ex}
\titlespacing*{\subsection}{0pt}{1.0ex plus .2ex minus .2ex}{0.45ex}
\titlespacing*{\subsubsection}{0pt}{1.0ex plus .2ex minus .2ex}{0.65ex}
\titlespacing*{\paragraph}{0pt}{0.65ex plus .15ex minus .1ex}{0.55em}
\usepackage{booktabs}
\usepackage{tabularx} 
\usepackage{caption}
\usepackage{subcaption} 

\usepackage{microtype} 
\usepackage[utf8]{inputenc}
\usepackage[T5]{fontenc}
\usepackage{csquotes}
\usepackage{tikzsymbols}

\usepackage{hyperref}

\usepackage{tgtermes}
\usepackage{newtxmath}

\usepackage{amsmath,amssymb}
\usepackage{graphicx}
\usepackage{booktabs}
\usepackage{tabularx}
\usepackage{array}
\usepackage{ragged2e}
\usepackage{cuted}

\newcolumntype{Y}{>{\RaggedRight\arraybackslash}X}
\newcolumntype{L}{
    >{\RaggedRight\arraybackslash}p{0.27\columnwidth}
}

\usepackage[section]{placeins} 

\usepackage{amsmath,amssymb}
\usepackage{graphicx}

\usepackage{xurl}   

\def\x{{\mathbf x}}
\def\L{{\cal L}}

\usepackage{url}
\usepackage[shortlabels]{enumitem}

\let\OLDthebibliography\thebibliography
\renewcommand\thebibliography[1]{
  \OLDthebibliography{#1}
  \setlength{\parskip}{0pt}
  \setlength{\itemsep}{0.10ex plus 0.10ex}
}

\begin{document}\sloppy
\topmargin = 0mm

\def\x{{\mathbf x}}
\def\L{{\cal L}}


\title{Direct Image-to-modern Vietnamese Translation of Han-Nom Manuscripts via MultiModal RLHF Preference Alignment}


\name{
Thi Kim Trang Vo$^{1,2}$,  Nghia Hieu Nguyen$^{1,2}$, Ha Minh Tan$^{1,2}$
}
\address{
$^{1}$ University of Information Technology, Ho Chi Minh City, Vietnam \\
$^{2}$ Vietnam National University, Ho Chi Minh City, Vietnam \\
trangvtk.18@grad.uit.edu.vn, nghiangh@uit.edu.vn, tanhm@uit.edu.vn. 
}

\maketitle

\begin{abstract}
Translating H\'an--N\^om manuscripts into modern Vietnamese is challenging
because historical pages are degraded, the script contains rare logographic
characters, and parallel supervision is limited. We propose a multimodal RLHF
preference-alignment framework that conditions Vietnamese generation on
manuscript images and aligned H\'an--N\^om source text. The model combines four
streams: CLIP ViT-L/14@336 for visual features,
\texttt{bert-base-chinese} for H\'an--N\^om representations,
\texttt{vinai/phobert-base} for Vietnamese representations, and T5-small encoder
states. Modality-specific projections and a fusion block compress the resulting
2{,}048-dimensional concatenation into a shared 512-dimensional
representation. Starting from the same supervised fine-tuned policy, we compare
PPO, DPO, and KTO under matched work-level macro-averaged evaluation. DPO
achieves the best BLEU-4, ROUGE-L, BERTScore, semantic similarity, CER, WER,
and token accuracy, whereas PPO obtains the highest precision, recall, and F1.
KTO remains competitive through its desirable--undesirable utility objective. All preference-aligned policies improve the BLEU-4 and semantic-similarity
scores available for the SFT baseline. These results indicate that multimodal
preference optimization complements supervised learning for lexical and
semantic quality in low-resource historical translation.
\end{abstract}


\section{Introduction}
\label{sec:introduction}

Historical Vietnamese manuscripts written in H\'an--N\^om preserve important
cultural, literary, and administrative knowledge, but remain difficult to access.
The script contains rare logographic characters, and surviving pages often
suffer from degradation, ink diffusion, irregular handwriting, and layout
variation. Parallel data linking manuscript images, H\'an--N\^om
transcriptions, and fluent modern Vietnamese translations are scarce. The task is
therefore not only OCR or text-to-text translation, but a coupled
image-to-language problem: the system must read degraded manuscript evidence,
recover source-script content, and generate faithful modern Vietnamese.

Prior work has mostly addressed this pipeline in separate stages. OCR resources
such as NomNaOCR~\cite{NomNaOCR}, IHR-NomDB~\cite{IHR-NomDB}, and the
Nom--Vietnamese Parallel Corpus~\cite{Nom-Vietnamese} support recognition,
transliteration, and translation research, while OCR systems such as
PaddleOCRv5~\cite{PaddleOCRv5} improve robustness on degraded pages. However,
OCR maps images to H\'an--N\^om text and does not directly generate fluent
Vietnamese. In parallel, SMT, NMT, and Transformer methods have been used for
N\^om-to-Quốc-ngữ, Chinese--Vietnamese, and H\'an--N\^om--Vietnamese
translation~\cite{SMT_Transliteration,Transformer_Translation_1,
Transformer_Translation_2,Translate_NMT}, but these methods usually assume clean
source text. LLM-based post-OCR correction~\cite{Thao_PostOCR} reduces
recognition noise, yet still operates after OCR and does not jointly align
visual evidence, source-script semantics, and target-language fluency.

Pure supervised Seq2Seq training is also limited for classical and literary
H\'an--N\^om. A passage may have multiple valid Vietnamese renderings: one may be
literal but unnatural, while another better preserves meaning, rhythm, and
literary nuance. Maximum-likelihood training imitates a single reference and
does not model preferences among alternatives, while BLEU-like metrics may miss
fluency, cultural adequacy, and style~\cite{Disadvantages_1,Disadvantages_2}.
These issues motivate a preference-based formulation.

We propose a multimodal RLHF framework for direct H\'an--N\^om manuscript
image-to-modern Vietnamese translation. Given a manuscript image, training-time
H\'an--N\^om supervision, and candidate Vietnamese outputs, the model learns
which output is more faithful, fluent, and stylistically appropriate.
Human-validated Vietnamese translations are used as chosen responses, while
controlled lower-quality variants are used as rejected responses. We first train
a LoRA-adapted T5-small policy by SFT, then align it with PPO~\cite{PPO},
DPO~\cite{DPO}, and KTO~\cite{KTO}. PPO uses reward/critic learning, DPO
optimizes chosen--rejected pairs directly, and KTO provides a utility-based
objective robust to noisy rejected samples.


Our contribution is not a new generic RLHF algorithm. Instead, we instantiate
preference optimization for low-resource multimodal historical translation. Our
contributions are: (i) formulating H\'an--N\^om manuscript translation as a
unified multimodal image-to-Vietnamese alignment problem; (ii) designing a
shared fusion backbone combining manuscript image, H\'an--N\^om, Vietnamese, and
T5 representations; (iii) constructing preference data with chosen references
and rejected variants for PPO, DPO, and KTO; and (iv) showing that preference
optimization complements SFT when maximum-likelihood learning is insufficient
for fluency, semantic fidelity, and literary style.

\paragraph{OCR usage.}

Our method is OCR-supervised but OCR-decoder-free at inference. H\'an--N\^om
transcriptions are used only during training to construct SFT and preference
data; the deployed system does not run a standalone OCR decoder or translate
from a separately decoded OCR output. Vietnamese generation is performed through
the multimodal policy from manuscript visual evidence and learned
representations.


\section{Methodology}
\label{sec:methodology}

\begin{figure*}[!t]
    \centering
    \includegraphics[width=0.94\textwidth]{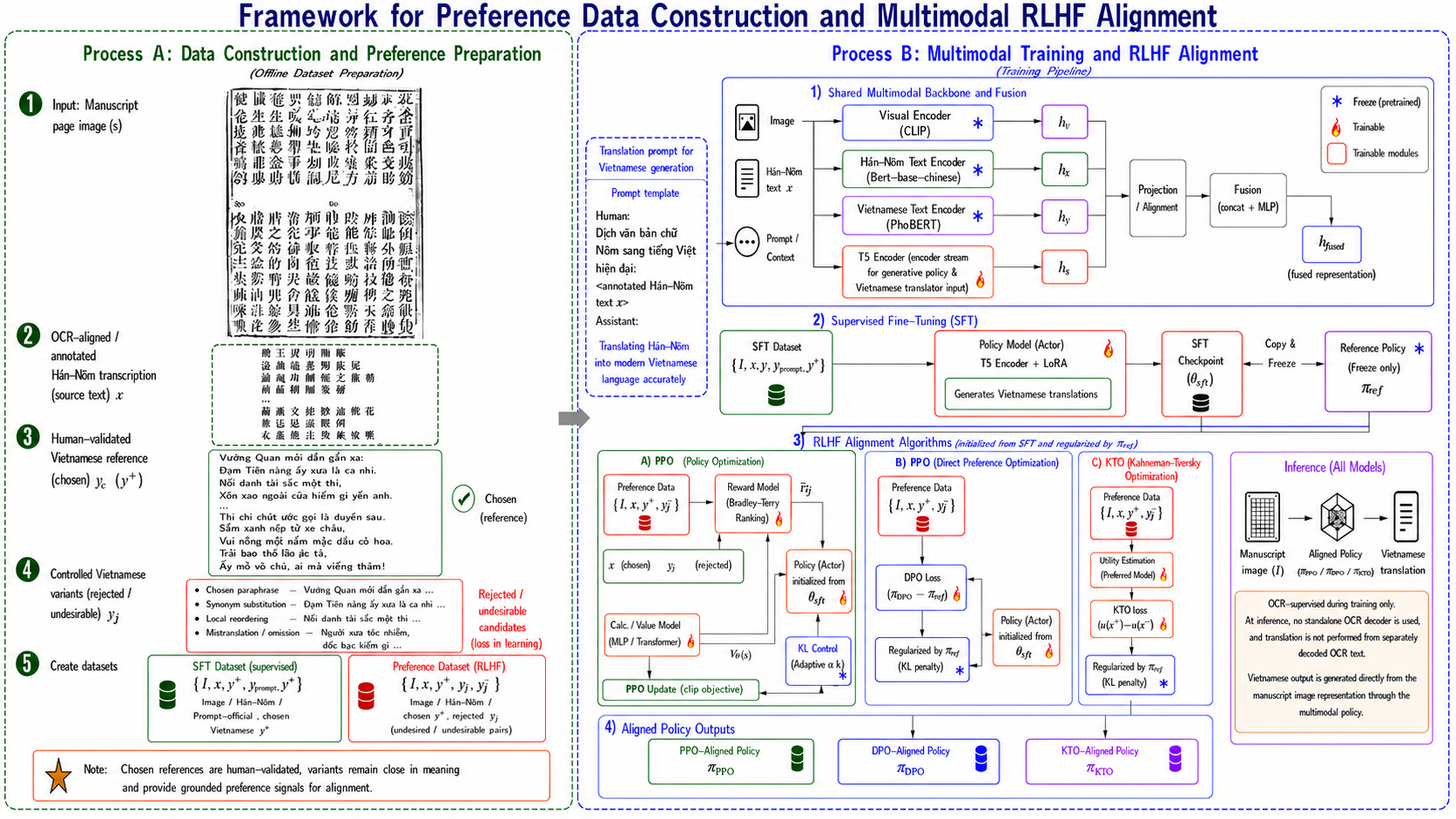}
    \caption{
        Framework for preference-data construction and multimodal RLHF
        alignment. PPO, DPO, and KTO share the same fusion backbone and
        LoRA-adapted T5 policy but use different preference objectives.
    }
    \label{fig:overall_framework}
\end{figure*}

\subsection{Data and Preference Construction}
\label{subsec:data_construction}

As illustrated in Fig.~\ref{fig:overall_framework}, we preprocess 2,953
Hán--Nôm manuscript pages from NomNaOCR~\cite{NomNaOCR} to construct
supervised translation and preference-alignment data. Each sample contains the
original manuscript image \(I\), an OCR-aligned or manually annotated
Hán--Nôm transcription \(x\), and a human-validated modern Vietnamese
translation \(y^{*}\). We validate each record, normalize Unicode and
whitespace, apply consistent character- or subword-level tokenization, and use
fuzzy matching to verify the corresponding \texttt{image\_path}.

The translation prompt is:

\begin{quote}
\footnotesize
\ttfamily
Human:\textbackslash n\\
Dịch văn bản chữ Nôm sang tiếng Việt hiện đại.\textbackslash n\\
(English: Translate the Hán--Nôm text accurately into modern Vietnamese.)%
\textbackslash n\\
\textless annotated Hán--Nôm text \(x\)\textgreater\textbackslash n\\
Assistant:\textbackslash n\\
\end{quote}

The verified translation is used as the chosen response \(y_c\). Rejected
responses \(y_r\) are generated through controlled paraphrasing, synonym
substitution, deletion, local reordering, mistranslation, or stylistic
weakening. These variants remain readable and semantically close to \(y_c\),
but are less faithful, complete, fluent, or stylistically appropriate. Each
sample is serialized into a unified JSON record containing
\texttt{prompt}, \texttt{input}, \texttt{output}, \texttt{chosen},
\texttt{rejected}, and \texttt{image\_path}, and is used for SFT, reward
modeling, PPO, DPO, and KTO alignment, which is directly consumed by downstream supervised fine-tuning (SFT) and RLHF stages.

\subsection{Multimodal Fusion and Supervised Fine-Tuning}
\label{subsec:fusion_sft}

The shared backbone in Fig.~\ref{fig:overall_framework} combines four streams:
CLIP ViT-L/14@336 visual features, Hán--Nôm representations from
\texttt{bert-base-chinese}, Vietnamese representations from
\texttt{vinai/phobert-base}, and T5-small encoder states. Let these streams be
\(H_I\), \(H_x\), \(H_v\), and \(H_t\).

Each stream is projected into a common 512-dimensional space:

\begin{equation}
\begin{aligned}
\widetilde{H}_I &= W_IH_I+b_I,
&
\widetilde{H}_x &= W_xH_x+b_x,
\\
\widetilde{H}_v &= W_vH_v+b_v,
&
\widetilde{H}_t &= W_tH_t+b_t.
\end{aligned}
\label{eq:fusion_projection}
\end{equation}

The projected features are concatenated as

\begin{equation}
\begin{aligned}
Z
&=
\left[
\widetilde{H}_I;
\widetilde{H}_x;
\widetilde{H}_v;
\widetilde{H}_t
\right],
\\
Z
&\in
\mathbb{R}^{L\times 2048}.
\end{aligned}
\label{eq:fusion_concat}
\end{equation}

The concatenated representation is compressed by a
linear--ReLU--dropout block:

\begin{equation}
\begin{aligned}
h_{\mathrm{fused}}
&=
\operatorname{Dropout}
\Bigl(
\operatorname{ReLU}(W_fZ+b_f)
\Bigr),
\\
h_{\mathrm{fused}}
&\in
\mathbb{R}^{L\times 512}.
\end{aligned}
\label{eq:fusion_output}
\end{equation}

A LoRA-adapted T5 decoder conditions on \(h_{\mathrm{fused}}\) to generate
Vietnamese. LoRA is applied to the query and value projections with rank
\(r=16\), scaling coefficient \(\alpha=16\), and dropout \(0.1\).

For the verified target
\(y^{*}=(y_1^{*},\ldots,y_T^{*})\), define

\begin{equation}
M_y=\sum_{t=1}^{T}m_t,
\label{eq:target_mask_count}
\end{equation}

where \(m_t\) masks padding and non-target positions. SFT minimizes

\begin{equation}
\begin{aligned}
\mathcal{L}_{\mathrm{SFT}}
&=
-\frac{1}{M_y}
\sum_{t=1}^{T}
m_t
\log
\pi_\theta
\left(
y_t^{*}
\mid
y_{<t}^{*},
h_{\mathrm{fused}}
\right).
\end{aligned}
\label{eq:sft_loss}
\end{equation}

The resulting checkpoint is copied and frozen as the reference policy:

\begin{equation}
\pi_{\mathrm{ref}}
=
\pi_{\theta_{\mathrm{SFT}}}.
\label{eq:reference_policy}
\end{equation}

\subsection{Multimodal Preference Alignment}
\label{subsec:preference_alignment}

We cast Hán--Nôm translation as offline multimodal RL. The state is the fused
representation conditioned on the manuscript image \(I\), aligned text \(x\),
and prompt \(p\):

\begin{equation}
s=h_{\mathrm{fused}}(I,x,p).
\label{eq:state_definition}
\end{equation}

The actor is the T5 decoder policy \(\pi_\theta\), which generates Vietnamese
tokens autoregressively:

\begin{equation}
\pi_\theta(y_t\mid y_{<t},s).
\label{eq:token_policy}
\end{equation}

The environment is a fixed dataset of chosen--rejected pairs:

\begin{equation}
\mathcal{D}_{\mathrm{pref}}
=
\{(I,x,p,y_c,y_r):y_c\succ y_r\},
\label{eq:offline_environment}
\end{equation}

where \(y_c\) and \(y_r\) denote the chosen and rejected translations. PPO uses
a learned reward model and critic, whereas DPO and KTO optimize policy
log-ratios relative to the frozen SFT reference.

\subsubsection{Proximal Policy Optimization (PPO)}
\label{subsubsec:ppo}

PPO~\cite{PPO} uses a reward model trained with Bradley--Terry ranking.
It independently fuses manuscript visual evidence, Hán--Nôm features, and
candidate Vietnamese representations to produce a scalar preference score:

\begin{equation}
\begin{aligned}
r_c &= r_\psi(I,x,y_c),\\
r_r &= r_\psi(I,x,y_r),\\
\mathcal{L}_{\mathrm{RM}}
&= -\log\sigma(r_c-r_r).
\end{aligned}
\label{eq:ppo_reward_model}
\end{equation}

The critic estimates the expected reward:

\begin{equation}
\begin{aligned}
V_{\phi}(s) &= W_{\phi}h_{\mathrm{fused}}^{\mathrm{pool}}+b_{\phi},\\
R &= r_{\psi}(I,x,y),\\
\mathcal{L}_{\mathrm{value}} &= \left(V_{\phi}(s)-R\right)^2.
\end{aligned}
\end{equation}


The token-level ratio, clipped ratio, and advantage are

\begin{equation}
\begin{aligned}
\rho_t
&=
\frac{
\pi_\theta(y_t\mid s_t)
}{
\pi_{\theta_{\mathrm{old}}}(y_t\mid s_t)
},
\\
\bar{\rho}_t
&=
\operatorname{clip}
\left(
\rho_t,
1-\epsilon,
1+\epsilon
\right),
\\
A_t
&=
R-V_\phi(s).
\end{aligned}
\label{eq:ppo_ratio_advantage}
\end{equation}

The clipped policy loss is

\begin{equation}
\begin{aligned}
\mathcal{L}_{\mathrm{clip}}
&=
-\frac{1}{M_y}
\sum_{t}
m_t
\min
\left\{
\rho_tA_t,
\bar{\rho}_tA_t
\right\}.
\end{aligned}
\label{eq:ppo_clip_loss}
\end{equation}

The complete PPO objective is

\begin{equation}
\begin{aligned}
\mathcal{L}_{\mathrm{PPO}}
&=
\mathcal{L}_{\mathrm{clip}}
+
\lambda_v
\mathcal{L}_{\mathrm{value}}
\\
&\quad
+
\lambda_{\mathrm{ref}}
\mathcal{L}_{\mathrm{ref}}
+
\lambda_{\mathrm{SFT}}
\mathcal{L}_{\mathrm{SFT}},
\end{aligned}
\label{eq:ppo_total_loss}
\end{equation}

where \(\mathcal{L}_{\mathrm{ref}}\) is a masked reference-anchor penalty based on the mean policy--reference token log-probability difference, rather than an exact theoretical KL divergence.


\subsubsection{Direct Preference Optimization (DPO)}
\label{subsubsec:dpo}

DPO~\cite{DPO} directly optimizes the chosen--rejected margin relative to the frozen
reference policy:

\begin{equation}
\begin{aligned}
\Delta_{\mathrm{DPO}}
&=
\log
\frac{
\pi_\theta(y_c\mid s)
}{
\pi_{\mathrm{ref}}(y_c\mid s)
}
\\
&\quad
-
\log
\frac{
\pi_\theta(y_r\mid s)
}{
\pi_{\mathrm{ref}}(y_r\mid s)
}.
\end{aligned}
\label{eq:dpo_margin}
\end{equation}

The DPO loss is

\begin{equation}
\mathcal{L}_{\mathrm{DPO}}
=
-\mathbb{E}
\left[
\log
\sigma
\left(
\beta
\Delta_{\mathrm{DPO}}
\right)
\right].
\label{eq:dpo_loss}
\end{equation}

DPO therefore requires neither a separate reward model nor a critic.

\subsubsection{Kahneman--Tversky Optimization (KTO)}
\label{subsubsec:kto}

KTO~\cite{KTO} learns from individually labeled desirable and undesirable responses. The
reference-normalized log-ratio is

\begin{equation}
\begin{aligned}
z_\theta(y\mid s)
&=
\log
\pi_\theta(y\mid s)
\\
&\quad
-
\log
\pi_{\mathrm{ref}}(y\mid s).
\end{aligned}
\label{eq:kto_log_ratio}
\end{equation}

Let \(\widehat{D}_{\mathrm{KL}}\) denote the detached mini-batch reference log-ratio baseline.
For desirable response \(y_D\),

\begin{equation}
\begin{aligned}
\ell_D
&=
1-
\sigma
\Bigl(
\beta
[
z_\theta(y_D\mid s)
-
\widehat{D}_{\mathrm{KL}}
]
\Bigr).
\end{aligned}
\label{eq:kto_desirable}
\end{equation}

For undesirable response \(y_U\),

\begin{equation}
\begin{aligned}
\ell_U
&=
1-
\sigma
\Bigl(
\beta
[
\widehat{D}_{\mathrm{KL}}
-
z_\theta(y_U\mid s)
]
\Bigr).
\end{aligned}
\label{eq:kto_undesirable}
\end{equation}

The KTO objective is

\begin{equation}
\mathcal{L}_{\mathrm{KTO}}
=
\mathbb{E}
\left[
\lambda_D\ell_D
+
\lambda_U\ell_U
\right].
\label{eq:kto_loss}
\end{equation}

KTO is suitable when undesirable responses are noisy, sparse, or incompletely
paired.

\subsection{Controlled Comparison and Inference}
\label{subsec:controlled_comparison}

{\small
PPO, DPO, and KTO share the same preprocessing, multimodal backbone,
LoRA-adapted policy, SFT initialization, frozen reference policy, and matched evaluation data. Their main differences lie in the alignment objectives and
algorithm-specific training modules: PPO uses reward and critic learning,
whereas DPO and KTO optimize reference-anchored preference losses. The contribution is therefore a multimodal
preference-alignment formulation for Hán--Nôm translation rather than a new
generic RLHF algorithm.

Hán--Nôm transcriptions provide training-time supervision for SFT and
preference construction. At inference, the deployed policy receives the
manuscript image and prompt and does not run a standalone OCR decoder or
translate from separately decoded OCR text.
}

%
%

\section{Experiments}
\label{sec:experiments}

\begin{table*}[!t]
\centering
\caption{Objective-level comparison under the same multimodal backbone, SFT
initialization, and matched evaluation protocol.}
\label{tab:objective_comparison}

\scriptsize
\setlength{\tabcolsep}{2.9pt}
\renewcommand{\arraystretch}{1.01}

\begin{tabularx}{\textwidth}{
@{}
lcccc
>{\RaggedRight\arraybackslash}p{0.20\textwidth}
Y
@{}}
\toprule
\textbf{Method}
& \textbf{Reward}
& \textbf{Critic}
& \textbf{Ref./KL}
& \textbf{SFT Init.}
& \textbf{Preference Form}
& \textbf{Observed Behavior}
\\
\midrule
SFT-only
& \(\times\)
& \(\times\)
& \(\times\)
& \(\checkmark\)
& Supervised target generation
& Frozen reference and preference-alignment baseline.
\\
PPO
& \(\checkmark\)
& \(\checkmark\)
& \(\checkmark\)
& \(\checkmark\)
& Reward-guided actor--critic update
& Best precision, recall, and F1; broader reference-content coverage.
\\
DPO
& \(\times^{\ast}\)
& \(\times^{\ast}\)
& \(\checkmark\)
& \(\checkmark\)
& Paired chosen--rejected optimization
& Best overlap, semantic, edit-distance, and token-accuracy metrics.
\\
KTO
& \(\times^{\ast}\)
& \(\times^{\ast}\)
& \(\checkmark\)
& \(\checkmark\)
& Desirable--undesirable utility
& Competitive performance under imperfect negative feedback.
\\
\bottomrule
\end{tabularx}

\vspace{0.03em}

\begin{minipage}{0.985\textwidth}
\scriptsize
\raggedright
\(\times^{\ast}\) indicates that reward and critic models are not core
components of the DPO or KTO objectives, although they may still be loaded for
scoring, diagnostics, or output selection. DPO uses chosen--rejected
log-probability ratios relative to the frozen SFT reference. KTO uses
desirable--undesirable utility with reference/KL control, whereas PPO directly
uses reward, critic, reference, and SFT-control terms.
\end{minipage}
\end{table*}

\subsection{Setup and Evaluation Protocol}
\label{subsec:experimental_setup}

{\small
Experiments use NomNaOCR dataset~\cite{NomNaOCR} and an implementation built on the environment of
OpenRLHF~\cite{OpenRLHF}, HuggingFace Transformers, and PEFT. Besides, SFT, PPO, DPO,
and KTO share a T5-small base model with approximately 60M parameters and the
same multimodal fusion backbone. CLIP ViT-L/14@336 for visual features,
\texttt{bert-base-chinese} for Hán--Nôm features,
\texttt{vinai/phobert-base} for Vietnamese features, and T5 encoder states are
fused into \(h_{\mathrm{fused}}\in\mathbb{R}^{L\times512}\). The dataset was split into training and validation sets at an 80:20 ratio, with 2{,}362 samples for training and 591 samples for validation. 
}

{\small
LoRA is applied to the T5 decoder query and value projections with
\(r=16\), \(\alpha=16\), and dropout \(0.1\), yielding approximately 1.2M
trainable parameters. Training uses AdamW with learning rate
\(2\times10^{-5}\), cosine scheduling, FP16 mixed-precision training, batch size \(1\), and maximum
sequence length \(256\). Due to the heavy com
putational cost, each variant is trained for a single epoch, which already requires more than 36 hours on one NVIDIA A100 GPU. PPO uses \(\epsilon=0.2\),
\(\lambda_v=0.1\), \(\lambda_{\mathrm{ref}}=0.02\), and
\(\lambda_{\mathrm{SFT}}=0.1\); DPO uses \(\beta=0.1\); and KTO uses
\(\beta=0.1\) with \(\lambda_D=\lambda_U=1.0\).
Table~\ref{tab:objective_comparison} summarizes the objective differences.
}

All aligned policies start from the same SFT checkpoint and are evaluated on
identical sample identities within each \texttt{work\_type}. For common sample
set \(\mathcal{I}_w\), method \(a\), and metric \(m\),

\begin{equation}
\begin{aligned}
\mathcal{I}^{\mathrm{PPO}}_w
&=
\mathcal{I}^{\mathrm{DPO}}_w
=
\mathcal{I}^{\mathrm{KTO}}_w
=
\mathcal{I}_w,
\\
\bar m_a
&=
\frac{1}{|\mathcal{W}|}
\sum_{w\in\mathcal{W}}
\frac{1}{|\mathcal{I}_w|}
\sum_{i\in\mathcal{I}_w}
m_{a,i}.
\end{aligned}
\label{eq:matched_macro_compact}
\end{equation}

The protocol first averages results within each manuscripts on the held-out validation
split and then macro-averages across works. This gives every work equal weight
and reduces bias from unequal sample coverage or manuscript length.

We report BLEU-4, ROUGE-L, BERTScore, semantic similarity, a KL-control
diagnostic, CER, WER, token accuracy, precision, recall, and F1. Let
\(y^*\) be the reference, \(\hat y\) the prediction,
\(d_{\mathrm{Lev}}\) the Levenshtein distance, and
\(M=\sum_u\min(c_{y^*}(u),c_{\hat y}(u))\) the multiset token overlap.
For compact notation, let
\(\mathcal{C}(y)=\operatorname{chars}(y)\) and
\(\mathcal{T}(y)=\operatorname{token}(y)\). The metrics are

\begin{equation}
\begin{aligned}
\mathrm{CER}
&=
\frac{
d_{\mathrm{Lev}}(\mathcal{C}(y^*),\mathcal{C}(\hat y))
}{
\max\{|\mathcal{C}(y^*)|,1\}
},
\\
\mathrm{WER}
&=
\frac{
d_{\mathrm{Lev}}(\mathcal{T}(y^*),\mathcal{T}(\hat y))
}{
\max\{|\mathcal{T}(y^*)|,1\}
},
\\
\mathrm{Precision}
&=
\frac{M}{\max\{|\mathcal{T}(\hat y)|,1\}},
\\
\mathrm{Recall}
&=
\frac{M}{\max\{|\mathcal{T}(y^*)|,1\}},
\\
\mathrm{F1}
&=
\frac{
2\,\mathrm{Precision}\,\mathrm{Recall}
}{
\max\{\mathrm{Precision}+\mathrm{Recall},\varepsilon\}
}.
\end{aligned}
\label{eq:evaluation_metrics}
\end{equation}

The implementation-level token accuracy is

\begin{equation}
\mathrm{Acc}_{\mathrm{tok}}
=
0.70F_{\mathrm{tok}}
+
0.25S_{\mathrm{seq}}
+
0.05S_{\mathrm{len}}.
\label{eq:token_accuracy_compact}
\end{equation}

BLEU-4 measures exact local \(n\)-gram agreement, ROUGE-L captures sequence
structure, BERTScore and semantic similarity evaluate contextual meaning,
CER/WER measure surface-form errors, and precision--recall metrics measure
token coverage. KL is used only as a reference-control diagnostic.

\subsection{Quantitative Results}
\label{subsec:quantitative_results}

Table~\ref{tab:macro_main_metrics} reports matched work-level
macro-averages using identical evaluation samples.

\begin{table}[!htbp]
\centering
\caption{Matched work-level macro-averages; higher is better except for KL,
CER, and WER.}
\label{tab:macro_main_metrics}

\fontsize{6.65}{7.20}\selectfont
\setlength{\tabcolsep}{1.45pt}
\renewcommand{\arraystretch}{1.04}

\begin{tabular*}{\columnwidth}{
@{\extracolsep{\fill}}lccccc@{}}
\toprule
\multicolumn{6}{c}{\textbf{A. Translation and semantic quality}}\\[-0.4mm]
\midrule
\textbf{Method}
& \textbf{BLEU-4\(\uparrow\)}
& \textbf{R-L\(\uparrow\)}
& \textbf{BERT\(\uparrow\)}
& \textbf{Sem.\(\uparrow\)}
& \textbf{KL\(\downarrow\)}
\\
\midrule
SFT-only
& 0.5085 & -- & -- & 0.8192 & --\\
PPO
& 0.6493 & 0.9239 & 0.8774 & 0.8890 & \textbf{1.99995}\\
DPO
& \textbf{0.6540}
& \textbf{0.9289}
& \textbf{0.8786}
& \textbf{0.8994}
& 2.00000\\
KTO
& 0.6459 & 0.9272 & 0.8761 & 0.8747 & 1.99997\\
\bottomrule
\end{tabular*}

\vspace{0.18em}

\begin{tabular*}{\columnwidth}{
@{\extracolsep{\fill}}lcccccc@{}}
\toprule
\multicolumn{7}{c}{\textbf{B. Surface error and token quality}}\\[-0.4mm]
\midrule
\textbf{Method}
& \textbf{CER\(\downarrow\)}
& \textbf{WER\(\downarrow\)}
& \textbf{TokAcc\(\uparrow\)}
& \textbf{Prec.\(\uparrow\)}
& \textbf{Rec.\(\uparrow\)}
& \textbf{F1\(\uparrow\)}
\\
\midrule
SFT-only
& -- & -- & -- & -- & -- & --\\
PPO
& 0.1133
& 0.1339
& 0.9332
& \textbf{0.8248}
& \textbf{0.7316}
& \textbf{0.7732}\\
DPO
& \textbf{0.1059}
& \textbf{0.1255}
& \textbf{0.9360}
& 0.8224
& 0.7274
& 0.7698\\
KTO
& 0.1082
& 0.1290
& 0.9345
& 0.8184
& 0.7217
& 0.7648\\
\bottomrule
\end{tabular*}

\vspace{0.12em}

\parbox{\columnwidth}{%
\fontsize{5.40}{5.90}\selectfont
\raggedright
R-L: ROUGE-L; Sem.: semantic similarity; TokAcc: token accuracy.
All policies use identical samples and work-first macro-averaging.
KL is a logged reference-control diagnostic, not an exact policy divergence.
}

\end{table}

\paragraph{Overall comparison.}
All preference-aligned policies improve the available SFT baseline. DPO obtains
the best BLEU-4, ROUGE-L, BERTScore, semantic similarity, CER, WER, and token
accuracy, consistent with direct optimization of fine-grained
chosen--rejected differences. PPO gives the highest precision, recall, and F1
because its sequence-level reward model and critic favor broader
reference-content recovery. KTO remains close to DPO and provides competitive
performance with pointwise desirable--undesirable feedback.

\paragraph{Gains over the SFT policy.}
Relative to SFT-only, PPO, DPO, and KTO improve BLEU-4 by $0.1408$, $0.1455$,
and $0.1374$, respectively, and semantic similarity by $0.0698$, $0.0802$,
and $0.0555$. The consistent direction of these gains supports the use of
preference alignment after supervised fine-tuning. Metrics unavailable for the
SFT checkpoint are left unreported rather than estimated, so comparisons for
ROUGE-L, BERTScore, CER, WER, and token-level quality are restricted to PPO,
DPO, and KTO.

\paragraph{Why ROUGE-L and BERTScore exceed BLEU-4.}
The \(0.8\)--\(0.9\) range mainly occurs for ROUGE-L, BERTScore, and semantic
similarity. Controlled Vietnamese variants may preserve the chosen reference
through synonyms, paraphrases, or local reordering. ROUGE-L rewards preserved
subsequence structure, and BERTScore uses contextual embeddings with soft token
alignment. BLEU-4 instead requires exact four-gram matches, so one substitution
or reordering breaks several neighboring \(n\)-grams. High ROUGE-L and
BERTScore with moderate BLEU-4 therefore reflects complementary structural,
semantic, and lexical evaluation rather than contradictory results.

The metric profile reflects each objective. DPO enlarges the
chosen--rejected log-probability margin relative to the SFT reference, yielding
stronger lexical, structural, and semantic accuracy. PPO optimizes a
sequence-level reward through an actor--critic framework, favoring broader
content coverage and higher precision, recall, and F1. KTO uses pointwise desirable--undesirable utility and is designed to
accommodate noisy or incompletely paired feedback, although its pairwise supervision is weaker than DPO. The small margins should be viewed as empirical trends rather than formal significance.

\begin{table*}[!t]
\centering
\caption{Contextual cross-model benchmarks for
Hán--Nôm-to-Vietnamese translation.}
\label{tab:cross_benchmark}

\setlength{\abovecaptionskip}{2pt}
\setlength{\belowcaptionskip}{1pt}

{\fontsize{5.15}{5.55}\selectfont
\setlength{\tabcolsep}{1.0pt}
\renewcommand{\arraystretch}{0.92}

\begin{tabular*}{\textwidth}{
@{\extracolsep{\fill}}lccccccccccc@{}}
\toprule
\multicolumn{12}{c}{
\textbf{B1. RLHF references and aligned policies}}\\[-0.45mm]
\midrule
\textbf{Model}
& \textbf{BLEU\(\uparrow\)}
& \textbf{R-L\(\uparrow\)}
& \textbf{BERT\(\uparrow\)}
& \textbf{Sem.\(\uparrow\)}
& \textbf{KL\(\downarrow\)}
& \textbf{CER\(\downarrow\)}
& \textbf{WER\(\downarrow\)}
& \textbf{TokAcc\(\uparrow\)}
& \textbf{Precision\(\uparrow\)}
& \textbf{Recall\(\uparrow\)}
& \textbf{F1\(\uparrow\)}
\\
\midrule
InstructGPT (PPO)~\cite{Instruct_GPT}
& 0.34 & 0.23 & -- & -- & \(\sim0.02\)
& -- & -- & -- & -- & -- & --\\

HH-RLHF (PPO)~\cite{Anthropic_HH-RLHF}
& -- & 0.85 & -- & -- & 0.015
& -- & -- & -- & -- & -- & --\\

Ours--SFT
& 0.5085 & -- & -- & 0.8192 & --
& -- & -- & -- & -- & -- & --\\

Ours--PPO
& 0.6493 & 0.9239 & 0.8774 & 0.8890 & 1.99995
& 0.1133 & 0.1339 & 0.9332
& \textbf{0.8248} & \textbf{0.7316} & \textbf{0.7732}\\

Ours--DPO
& \textbf{0.6540}
& \textbf{0.9289}
& \textbf{0.8786}
& \textbf{0.8994}
& 2.00000
& \textbf{0.1059}
& \textbf{0.1255}
& \textbf{0.9360}
& 0.8224 & 0.7274 & 0.7698\\

Ours--KTO
& 0.6459 & 0.9272 & 0.8761 & 0.8747 & 1.99997
& 0.1082 & 0.1290 & 0.9345
& 0.8184 & 0.7217 & 0.7648\\
\bottomrule
\end{tabular*}
}

\vspace{0.02em}

{\fontsize{5.25}{5.65}\selectfont
\setlength{\tabcolsep}{1.2pt}
\renewcommand{\arraystretch}{0.91}

\begin{tabularx}{\textwidth}{
@{}
>{\RaggedRight\arraybackslash}p{0.40\textwidth}
>{\centering\arraybackslash}p{0.08\textwidth}
>{\centering\arraybackslash}p{0.08\textwidth}
Y
@{}}
\toprule
\multicolumn{4}{c}{
\textbf{B2. Domain text and transliteration references}}\\[-0.45mm]
\midrule
\textbf{Model}
& \textbf{BLEU\(\uparrow\)}
& \textbf{CER\(\downarrow\)}
& \textbf{Direction / Setting}
\\
\midrule
SMT transliteration~\cite{SMT_Transliteration}
& 0.857
& --
& Text-only Nôm-to-Vietnamese.
\\

Multilingual NMT~\cite{NMT_Transliteration}
& \(+6.3^{\dagger}\)
& --
& Text-only multilingual transliteration.
\\

T5 baseline~\cite{Duong2026SinoNom}
& 0.3874
& 0.38
& Vietnamese-to-Sino-Nôm reverse direction.
\\

Computational Linguistics Center transliteration tool
~\cite{Duong2026SinoNom}
& 0.3883
& 0.38
& Vietnamese-to-Sino-Nôm reverse direction.
\\

Dual-branch T5~\cite{Duong2026SinoNom}
& 0.6973
& 0.16
& Vietnamese-to-Sino-Nôm reverse direction.
\\
\bottomrule
\end{tabularx}
}

\vspace{0.02em}

{\fontsize{5.25}{5.65}\selectfont
\setlength{\tabcolsep}{1.2pt}
\renewcommand{\arraystretch}{0.91}

\begin{tabularx}{\textwidth}{
@{}
>{\RaggedRight\arraybackslash}p{0.40\textwidth}
>{\centering\arraybackslash}p{0.08\textwidth}
>{\centering\arraybackslash}p{0.08\textwidth}
Y
@{}}
\toprule
\multicolumn{4}{c}{
\textbf{B3. OCR-like output-error references}}\\[-0.45mm]
\midrule
\textbf{Method}
& \textbf{CER\(\downarrow\)}
& \textbf{WER\(\downarrow\)}
& \textbf{Setting}
\\
\midrule
GPT-4o~\cite{Thao_PostOCR}
& 0.05 & 0.11 & Standalone/post-OCR reference.\\

Azure Document AI~\cite{Thao_PostOCR}
& 0.04 & 0.13 & Standalone/post-OCR reference.\\

Google Document AI~\cite{Thao_PostOCR}
& 0.07 & 0.14 & Standalone/post-OCR reference.\\

Fine-tuned Tesseract~\cite{Thao_PostOCR}
& 0.09 & 0.18 & Standalone/post-OCR reference.\\
\bottomrule
\end{tabularx}
}

\vspace{0.01em}

\begin{minipage}{\textwidth}
\fontsize{4.75}{5.15}\selectfont
\raggedright
R-L denotes ROUGE-L; Sem. denotes semantic similarity; TokAcc denotes the
code-level token accuracy. External baselines report only the metrics available
in their original settings. \(^{\dagger}\Delta\mathrm{BLEU}=+6.3\) denotes
improvement over the corresponding multilingual NMT baseline; its absolute BLEU
score was not reported. External transliteration systems use different
directions or text-only inputs and therefore serve as contextual references
rather than direct competitors.
\end{minipage}

\vspace{-0.25em}
\end{table*}

\begin{table}[!t]
\centering
\caption{Qualitative preference example from \textit{Truyện Kiều}
(\textit{Tale of Kiều}, 1872).}
\label{tab:demo_kieu_1872}

\fontsize{6.0}{6.55}\selectfont
\setlength{\tabcolsep}{2.0pt}
\renewcommand{\arraystretch}{1.0}

\begin{tabularx}{\columnwidth}{
@{}
>{\RaggedRight\arraybackslash}p{0.26\columnwidth}
>{\RaggedRight\arraybackslash}X
@{}}
\toprule
\textbf{Component}
&
\multicolumn{1}{c}{\textbf{Content}}
\\
\midrule

\begin{minipage}[t][0.150\textheight][b]{\linewidth}
\centering
\includegraphics[
width=0.92\linewidth,
height=0.105\textheight,
keepaspectratio
]{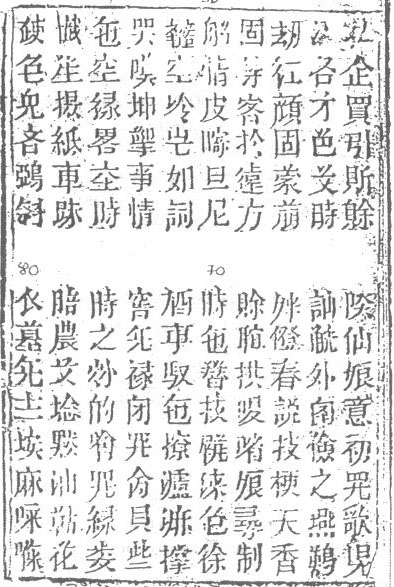}\\[-0.5mm]
\textbf{Manuscript}
\end{minipage}
&
\begin{minipage}[t][0.150\textheight][b]{\linewidth}
\centering
\includegraphics[
width=0.94\linewidth,
height=0.140\textheight,
keepaspectratio,
trim=24 8 24 8,
clip
]{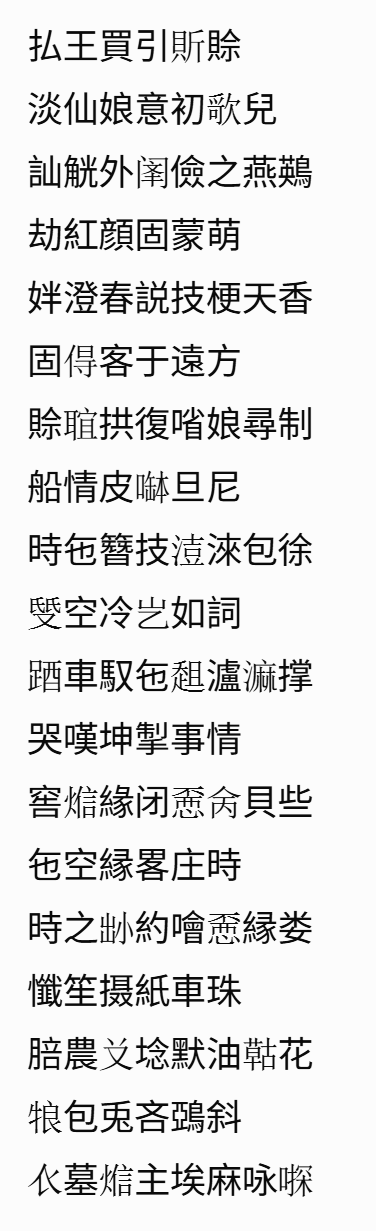}\\[-0.5mm]
\textbf{Aligned Nôm}
\end{minipage}
\\
\midrule

\textbf{Vietnamese reference}
&
\begin{minipage}[t]{\linewidth}
\vspace{0pt}
\RaggedRight
Vương Quan mới dẫn gần xa: \\
Đạm Tiên nàng ấy xưa là ca nhi. \\
Nổi danh tài sắc một thì, \\
Xôn xao ngoài cửa hiếm gì yến anh. \\
\textit{\ldots} \\
Thì chi chút ước gọi là duyên sau. \\
Sắm xanh nếp tử xe châu, \\
Vùi nông một nấm mặc dầu cỏ hoa. \\
Trải bao thỏ lặn ác tà, \\
Ấy mồ vô chủ, ai mà viếng thăm!
\end{minipage}
\\

\textbf{English rendering}
&
Vương Quan explains that Đạm Tiên was once a renowned singer whose beauty and
talent attracted many admirers; her grave later became unattended.
\\

\textbf{Controlled variants}
&
\textbf{V1:} close paraphrase;
\textbf{V2:} synonym and local-order variation preserving meaning and rhythm;
\textbf{V3:} stronger imagery and stylistic reformulation.
\\
\bottomrule
\end{tabularx}

\end{table}

\subsection{Contextual and Qualitative Analysis}
\label{subsec:contextual_qualitative}

Table~\ref{tab:cross_benchmark} provides contextual evidence rather than a
direct SOTA ranking. General RLHF systems address instruction following or
dialogue; transliteration systems assume clean symbolic text or evaluate the
reverse Vietnamese-to-Sino-Nôm direction; and OCR references perform
recognition or post-OCR correction. These systems differ from our setting in
input modality, output space, dataset, and metric definition. Their numerical
values therefore should not be interpreted as direct rankings against degraded
manuscript-image-to-modern-Vietnamese translation. In particular, our CER and
WER quantify surface distance between the generated Vietnamese translation and
the human-validated Vietnamese reference, rather than the character-recognition
accuracy of an independent OCR decoder. The principal evidence for comparing
PPO, DPO, and KTO is consequently the matched shared-backbone evaluation in
Table~\ref{tab:macro_main_metrics}.

Table~\ref{tab:demo_kieu_1872} illustrates the construction and difficulty of
the preference data. The human-validated Vietnamese text is the chosen
response. V1 introduces only limited paraphrasing; V2 applies synonym
substitution and local reordering while preserving meaning and poetic rhythm;
and V3 permits stronger imagery and stylistic reformulation. These candidates
remain plausible and source-related, making the preference task more
challenging than separating a valid translation from random or visibly
corrupted text.

The example shows why the metrics should be interpreted jointly. Although the
variants are generated for the RLHF setting, they are constrained to remain
semantically close to the human-validated chosen response, which serves as the
ground-truth translation, rather than introducing arbitrary or unrelated
outputs. Valid paraphrases may lower BLEU-4 because synonyms or local
reordering break exact four-grams, while ROUGE-L remains high when long
subsequences are preserved. BERTScore can also stay high by aligning
semantically related words despite surface differences. In contrast, omissions,
mistranslations, or unsupported additions tend to reduce semantic similarity,
recall, and ROUGE-L. The example therefore supports preference learning among
plausible Vietnamese renderings rather than exact reproduction of one
reference. It reflects preference-data difficulty, not a blinded comparison
among PPO, DPO, and KTO; expert assessment is still needed for historical
faithfulness, readability, poetic adequacy, and literary style.

\footnotesize{

\section{Conclusion and Future Work}
\label{sec:conclusion_future}

We presented a multimodal RLHF framework for direct H\'an--N\^om manuscript
image-to-Vietnamese translation that combines visual evidence, source-script
supervision, and Vietnamese generation while remaining OCR-decoder-free at
inference. Under the same backbone and matched work-level evaluation, DPO
achieves the best translation, semantic, edit-distance, and token-accuracy
metrics; PPO gives the highest precision, recall, and F1; and KTO remains competitive through its pointwise desirable--undesirable utility objective. These results show that
preference optimization complements SFT for semantic fidelity, lexical
faithfulness, and literary variation. The main limitation is the reliance on
curated NomNaOCR~\cite{NomNaOCR} samples with aligned H\'an--N\^om text and
verified Vietnamese references. Open-domain manuscripts from Google or digital
archives may lack both annotations and ground-truth translations. Future work
will address noisier collections through lightweight VLM-based text-region
extraction, domain-specific distillation with pseudo-labels or confidence-guided
signals, and expert evaluation of historical and literary quality.
}

\footnotesize\noindent
\textbf{Acknowledgment:} This research was funded by University of Information Technology, Vietnam National University Ho Chi Minh City under grant number CS1-2026-80583.


\vspace{-0.8em}
{\fontsize{7.35pt}{7.85pt}\selectfont
\bibliographystyle{IEEEtran}
\bibliography{ref}

@inproceedings{NomNaOCR,
  author    = {Dang, H.-Q. and Nguyen, D.-A. and Pham, P.-P. and others},
  title     = {{NomNaOCR}: The First Dataset for Optical Character Recognition on {Han--Nom} Script},
  booktitle = {Proc. RIVF},
  year      = {2022},
  pages     = {476--481},
  doi       = {10.1109/RIVF55975.2022.10013842}
}

@inproceedings{IHR-NomDB,
  author    = {Vu, M.-T. and Le, V. L. and Beurton-Aimar, M.},
  title     = {{IHR-NomDB}: The Old Degraded Vietnamese Handwritten Script Archive Database},
  booktitle = {Proc. ICDAR},
  year      = {2021},
  pages     = {85--99},
  doi       = {10.1007/978-3-030-86334-0_6}
}

@inproceedings{Nom-Vietnamese,
  author    = {Dang, M.-N. and Cao, T.-H. and Lam, T.-D. and others},
  title     = {Parallel Corpus Construction for Chinese and Vietnamese in Historical Texts},
  booktitle = {LNCS},
  year      = {2025},
  pages     = {189--203},
  doi       = {10.1007/978-3-031-98164-7_15}
}

@misc{PaddleOCRv5,
  author       = {Nguyen, M. H. and Thiet, S. N.},
  title        = {Enhancing {OCR} for Sino-Vietnamese Language Processing via Fine-tuned {PaddleOCRv5}},
  year         = {2025},
  howpublished = {arXiv:2510.04003}
}

@inproceedings{Thao_PostOCR,
  author    = {Do, T. and Tran, D. P. and Vo, A. and Kim, D.},
  title     = {Reference-Based Post-OCR Processing with {LLM} for Precise Diacritic Text in Historical Document Recognition},
  booktitle = {Proc. AAAI},
  year      = {2025},
  volume    = {39},
  number    = {27},
  pages     = {27951--27959}
}

@inproceedings{Translate_NMT,
  author    = {Chau, T. and Ngo, D.-V. and Nguyen, M.-T. and others},
  title     = {Translate {Han--Nom} to Vietnamese using Neural Machine Translation Methods},
  booktitle = {Proc. ATC},
  year      = {2023},
  pages     = {90--94},
  doi       = {10.1109/ATC58710.2023.10318891}
}

@article{SMT_Transliteration,
  author  = {Dinh, D. and Nguyen, P. and Nguyen, L. H. B.},
  title   = {Transliterating N{\^o}m Scripts into Vietnamese National Scripts using Statistical Machine Translation},
  journal = {IJACSA},
  year    = {2021},
  volume  = {12},
  number  = {2},
  doi     = {10.14569/IJACSA.2021.0120205}
}

@inproceedings{NMT_Transliteration,
  author    = {Hung, P. and Nguyen, L. and Dinh, D.},
  title     = {Transliterating Nom Script into Vietnamese National Script Using Multilingual Neural Machine Translation},
  booktitle = {Intelligent Systems and Data Science},
  year      = {2024},
  pages     = {133--147}
}

@misc{Transformer_Translation_1,
  author       = {Trieu, H. L. and Bui, S. K. and Tran, T. M. and others},
  title        = {{VBD-MT} Chinese-Vietnamese Translation Systems for {VLSP} 2022},
  year         = {2023},
  howpublished = {arXiv:2308.07601}
}

@misc{Transformer_Translation_2,
  author       = {Nguyen, P. M. and Nguyen, L. M.},
  title        = {An Effective Method using Phrase Mechanism in Neural Machine Translation},
  year         = {2023},
  howpublished = {arXiv:2308.10482}
}

@phdthesis{Disadvantages_1,
  author = {Zhou, F.},
  title  = {The Comparison of Translationese in Machine Translation and Human Translation in Terms of Translation Relations},
  school = {Ph.D. thesis},
  year   = {2022}
}

@misc{Disadvantages_2,
  author       = {Wieting, J. and Berg-Kirkpatrick, T. and Gimpel, K. and Neubig, G.},
  title        = {Beyond {BLEU}: Training Neural Machine Translation with Semantic Similarity},
  year         = {2019},
  howpublished = {arXiv:1909.06694}
}

@misc{PPO,
  author       = {Schulman, J. and Wolski, F. and Dhariwal, P. and Radford, A. and Klimov, O.},
  title        = {Proximal Policy Optimization Algorithms},
  year         = {2017},
  howpublished = {arXiv:1707.06347}
}

@misc{DPO,
  author       = {Rafailov, R. and Sharma, A. and Mitchell, E. and others},
  title        = {Direct Preference Optimization: Your Language Model is Secretly a Reward Model},
  year         = {2024},
  howpublished = {arXiv:2305.18290}
}

@misc{KTO,
  author       = {Ethayarajh, K. and Xu, W. and Muennighoff, N. and others},
  title        = {{KTO}: Model Alignment as Prospect Theoretic Optimization},
  year         = {2024},
  howpublished = {arXiv:2402.01306}
}

@misc{OpenRLHF,
  author       = {Hu, J. and Wu, X. and Shen, W. and others},
  title        = {{OpenRLHF}: An Easy-to-use, Scalable and High-performance {RLHF} Framework},
  year         = {2025},
  howpublished = {arXiv:2405.11143}
}

@misc{Instruct_GPT,
  author       = {Ouyang, L. and Wu, J. and Jiang, X. and others},
  title        = {Training Language Models to Follow Instructions with Human Feedback},
  year         = {2022},
  howpublished = {arXiv:2203.02155}
}

@misc{Anthropic_HH-RLHF,
  author       = {Bai, Y. and Jones, A. and Ndousse, K. and others},
  title        = {Training a Helpful and Harmless Assistant with Reinforcement Learning from Human Feedback},
  year         = {2022},
  howpublished = {arXiv:2204.05862}
}

@inproceedings{Duong2026SinoNom,
  author    = {Duong, Trieu and Dinh, Si-Dien and Tran, Anh and Nguyen, Tan and Ngo, Minh and Nguyen, Long},
  title     = {Transliteration of Vietnamese National Scripts into Sino-Nom Scripts Using Transformer-based Model},
  booktitle = {Computational Intelligence in Engineering Science},
  year      = {2026},
  pages     = {133--145},
  doi       = {10.1007/978-3-032-21625-0_10}
}
}


\end{document}